\documentclass[10pt,twocolumn,letterpaper]{article}

\usepackage{cvpr}
\usepackage{times}
\usepackage{epsfig}
\usepackage{epstopdf}
\usepackage{graphicx}
\usepackage{amsmath}
\usepackage{amssymb}
\usepackage{units}
\usepackage[printonlyused]{acronym}

\usepackage[pagebackref=true,breaklinks=true,letterpaper=true,colorlinks,bookmarks=false]{hyperref}
\usepackage{cleveref}

\acrodef{SfM}{Structure from Motion}

\cvprfinalcopy 

\def\cvprPaperID{3124} 

\def\b#1{\boldsymbol{#1}}

\ifcvprfinal\fi
\begin{document}

\title{CodeSLAM --- Learning a Compact, Optimisable Representation for Dense Visual SLAM}

\author{Michael Bloesch, Jan Czarnowski, Ronald Clark, Stefan Leutenegger, Andrew J. Davison \\
Dyson Robotics Laboratory at Imperial College, Department of Computing, Imperial College London, UK \\
{\tt\small m.bloesch@imperial.ac.uk}
}

\maketitle
\ifcvprfinal\thispagestyle{empty}\fi

\begin{abstract}
The representation of geometry in real-time 3D perception systems continues to be a critical research issue.
Dense maps capture complete surface shape and can be augmented with semantic labels, but their high dimensionality makes them computationally costly to store and process, and unsuitable for rigorous probabilistic inference.
Sparse feature-based representations avoid these problems, but capture only partial scene information and are mainly useful for localisation only.

We present a new compact but dense representation of scene geometry which is conditioned on the intensity data from a single image and generated from a code consisting of a small number of parameters.
We are inspired by work both on learned depth from images, and auto-encoders.
Our approach is suitable for use in a keyframe-based monocular dense SLAM system:
While each keyframe with a code can produce a depth map, the code can be optimised efficiently \emph{jointly} with pose variables and together with the codes of overlapping keyframes to attain global consistency.
Conditioning the depth map on the image allows the code to only represent aspects of the local geometry which cannot directly be predicted from the image.
We explain how to learn our code representation, and demonstrate its advantageous properties in monocular SLAM.
\end{abstract}

\section{Introduction}
The underlying representation of scene geometry is a crucial element of any localisation and mapping algorithm.
Not only does it influence the type of geometric qualities that can be mapped, but also dictates what algorithms can be applied.
In SLAM in general, but especially in monocular vision, where scene geometry cannot be retrieved from a single view, the representation of geometrical uncertainties is essential.
However, uncertainty propagation quickly becomes intractable for large degrees of freedom.
This difficulty has split mainstream SLAM approaches into two categories: 
\emph{sparse} SLAM ~\cite{Davison:ICCV2003,Klein:Murray:ISMAR2007,Mur-Artal:etal:TRO2015} which represents geometry by a sparse set of features and thereby allows joint probabilistic inference of structure and motion (which is a key pillar of probabilistic SLAM \cite{Durrant-Whyte:etal:RAM2006}) and \emph{dense} or {semi-dense} SLAM ~\cite{Newcombe:etal:ICCV2011,Engel:etal:ECCV2014} that attempts to retrieve a more complete description of the environment at the cost of approximations to the inference methods (often discarding cross-correlation of the estimated quantities and relying on alternating optimisation of pose and map~\cite{Platinsky:etal:ICRA2017,Engel:etal:PAMI2017}).

\begin{figure}[t]
  \begin{center}
    \includegraphics[width=1.0\linewidth]{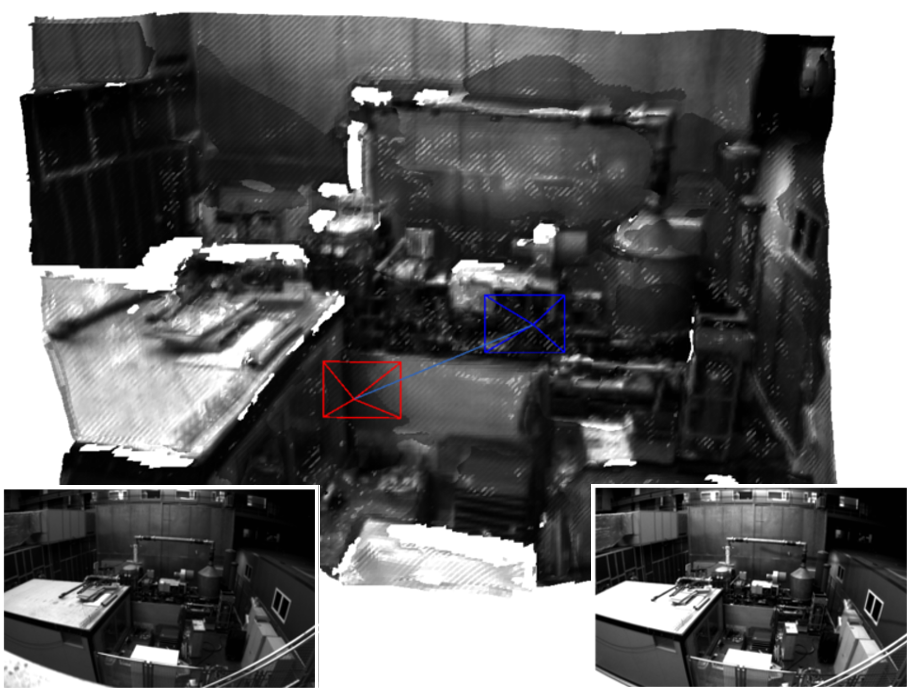}
  \end{center}
  \caption{
    Two view reconstruction on selected frames from the EuRoC dataset.
    The proposed compact representation of 3D geometry enables joint optimisation of the scene structure and relative camera motion without explicit priors and in near real-time performance.
  }
  \label{fig:sfm2_euroc}
  \vspace{2mm}\hrule
\end{figure}

However, the conclusion that a dense representation of the environment requires a large number of parameters is not necessarily correct.
The geometry of natural scenes is not a random collection of occupied and unoccupied space but exhibits a high degree of order.
In a depth map, the values of neighbouring pixels are highly correlated and can often be accurately represented by well known geometric smoothness primitives.
But more strongly, if a higher level of understanding is available, a scene could be decomposed into a set of semantic objects (e.g.\ a chair) together with some internal parameters (e.g.\ size of chair, number of legs) and a pose, following a direction indicated by the SLAM++ system~\cite{Salas-Moreno:etal:CVPR2013} towards representation with very few parameters.
Other more general scene elements which exhibit simple regularity such as planes can be recognised and efficiently parametrised within SLAM systems (e.g.~\cite{Salas-Moreno:etal:ISMAR2014,Kaess:ICRA2015}).
However, such human-designed dense abstractions are limited in the fraction of natural, cluttered scenes which they can represent.

In this work we aim at a more generic compact representation of dense scene geometry by training an auto-encoder on depth images.
While a straightforward auto-encoder might over-simplify the reconstruction of natural scenes, our key novelty is to condition the training on intensity images.
Our approach is planned to fit within the common and highly scalable keyframe-based SLAM paradigm~\cite{Klein:Murray:ISMAR2007,Engel:etal:ECCV2014}, where a scene map consists of a set of selected and estimated historical camera poses together with the corresponding captured images and supplementary local information such as depth estimates.
The intensity images are usually required for additional tasks, such as descriptor matching for place recognition or visualisation, and are thus readily available for supporting the depth encoding.

The depth map estimate for a keyframe thus becomes a function of the corresponding intensity image and an unknown compact representation (henceforth referred to as `code').
This allows for a compact representation of depth without sacrificing reconstruction detail.
In inference algorithms the code can be used as dense representation of the geometry and, due to its limited size, this allows for full joint estimation of both camera poses and dense depth maps for multiple overlapping keyframes.
We might think of the image providing local details and the code as supplying more global shape parameters which are often not predicted well by `depth from single image' learning.
Importantly though, these global shape parameters are not a designed geometric warp but have a learned space which tends to relate to semantic entities in the scene, and could be seen as a step towards enabling optimisation in general semantic space.

Our work comes at a time when many authors are combining techniques from deep learning with estimation-based SLAM frameworks, and there is an enormously fertile field of possibilities for this.
Some particularly eye-catching pieces of work over the past year have focused on supervised and self-supervised training of surprisingly capable networks which are able to estimate visual odometry, depth and other quantities from video~\cite{Garg:etal:ECCV2016,Ummenhofer:etal:ARXIV2016,Cadena:etal:RSS2016,Wang:etal:ICRA2017,Clark:etal:AAAI2017,Zhou:etal:CVPR2017,Yin:Shi:CVPR2018}.
These methods run with pure feed forward network operation at runtime, but rely on geometric and photometric formulation and understanding at training time to correctly formulate the loss functions which connect different network components.
Other systems are looking towards making consistent long-term maps and for instance combine learned normal predictions with photometric constraints at test time \cite{Weerasekera:etal:ICRA2017}.
Such systems are able to refine geometric estimates, and this is the domain in which we are particularly interested here.
In CNN-SLAM~\cite{Tateno:etal:CVPR2017} single image depth prediction and dense alignment are used to produce a dense 3D map and this gives a promising result, but it is not possible to optimise the predicted depth maps further for consistency when multiple keyframes overlap as it is in our approach.

To summarise, the two key contributions of our paper are:
\begin{itemize}
 \item The derivation of a compact and optimisable representation of dense geometry by conditioning a depth auto-encoder on intensity images.
 \item The implementation of the first real-time targeted monocular system that achieves such a tight joint optimisation of motion and dense geometry.
\end{itemize}
In the rest of this paper, we will first explain our method for depth learning and prediction, and then show the applicability of this approach in a SLAM setting.

\section{Intensity Conditioned Depth Auto-Encoding}
Two important qualities of geometry representations are \emph{accuracy} and \emph{practicality}.
While the \emph{accuracy} of a representation simply relates to its ability to reproduce the geometry, the \emph{practicality} describes how well the representation can be used in an overall system.
For inference-based SLAM systems, the latter typically requires the representation to lead to an optimisable loss function.
For a representation $G$ of the geometry a loss function $L(G)$ should be differentiable and have a clear minimum.
Additionally, the size of the representation $G$ should be limited in order to allow the estimation of second-order statistical moments (a covariance matrix) as part of more powerful inference methods.

\begin{figure}[t]
  \begin{center}
    \small
    Reconstruction \hspace{23mm} Groundtruth \hspace{4mm} \\[1mm]
    \includegraphics[width=1.0\linewidth]{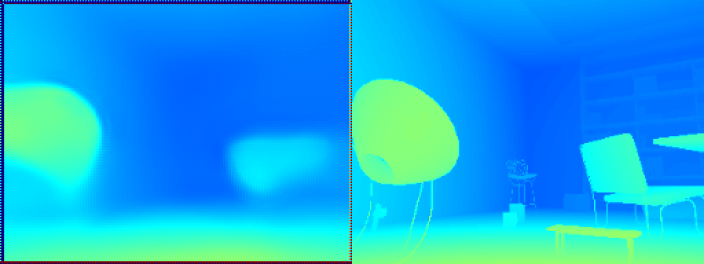}
  \end{center}
  \caption{
    Depth auto-encoder without the use of image intensity data.
    Due to the bottleneck of the auto-encoder only major traits of the depth image can be captured.
  }
  \label{fig:no_img}
  \vspace{2mm}\hrule
\end{figure}

In order to come up with a compact representation of the scene geometry we explore auto-encoder-like network architectures.
Auto-encoders are networks which attempt to learn an identity mapping while being subject to an information bottleneck which forces the network to find a compact representation of the data~\cite{Rumelhart:1986}.
In a naive attempt to auto-encode depth this would lead to very blurry depth reconstruction since only the major traits of the depth image can make it through the bottleneck (see \Cref{fig:no_img}).
In a monocular vision setup, however, we have access to the intensity images, which we are very likely to store alongside every keyframe.
This can be leveraged to make the encoding more efficient:
We do not need to encode the full depth information, but only need to retain the part of the information which cannot be retrieved from the intensities.
The depth $D$ thus becomes a function of image $I$ and (unknown) code $\b{c}$:
\begin{align}
  D = D(I, \b{c})~.
\end{align}

The above equation also highlights the relation to depth-from-mono architectures~\cite{Eigen:etal:NIPS2014, Liu:etal:2015, Garg:etal:ECCV2016, Zhou:etal:CVPR2017} which solve a code-less version of the problem, $D = D(I)$.
Essentially, the employed architecture is a combination of the depth-from-mono-architecture of Zhou et al.~\cite{Zhou:etal:CVPR2017} and a variational auto-encoder for depth.
We have chosen a variational auto-encoder network~\cite{Kingma:Welling:ICLR2014} in order to increase the smoothness of the mapping between code and depth: small changes in the code should lead to small changes in the depth.
While the \emph{practicality} of our representation is thus addressed by the smoothness and the limited code size, the \emph{accuracy} is maximised by training for the reconstruction error.

\subsection{Detailed Network Architecture}\label{sec:det_net}

\begin{figure}[b]
  \begin{center}
    \includegraphics[width=1.0\linewidth]{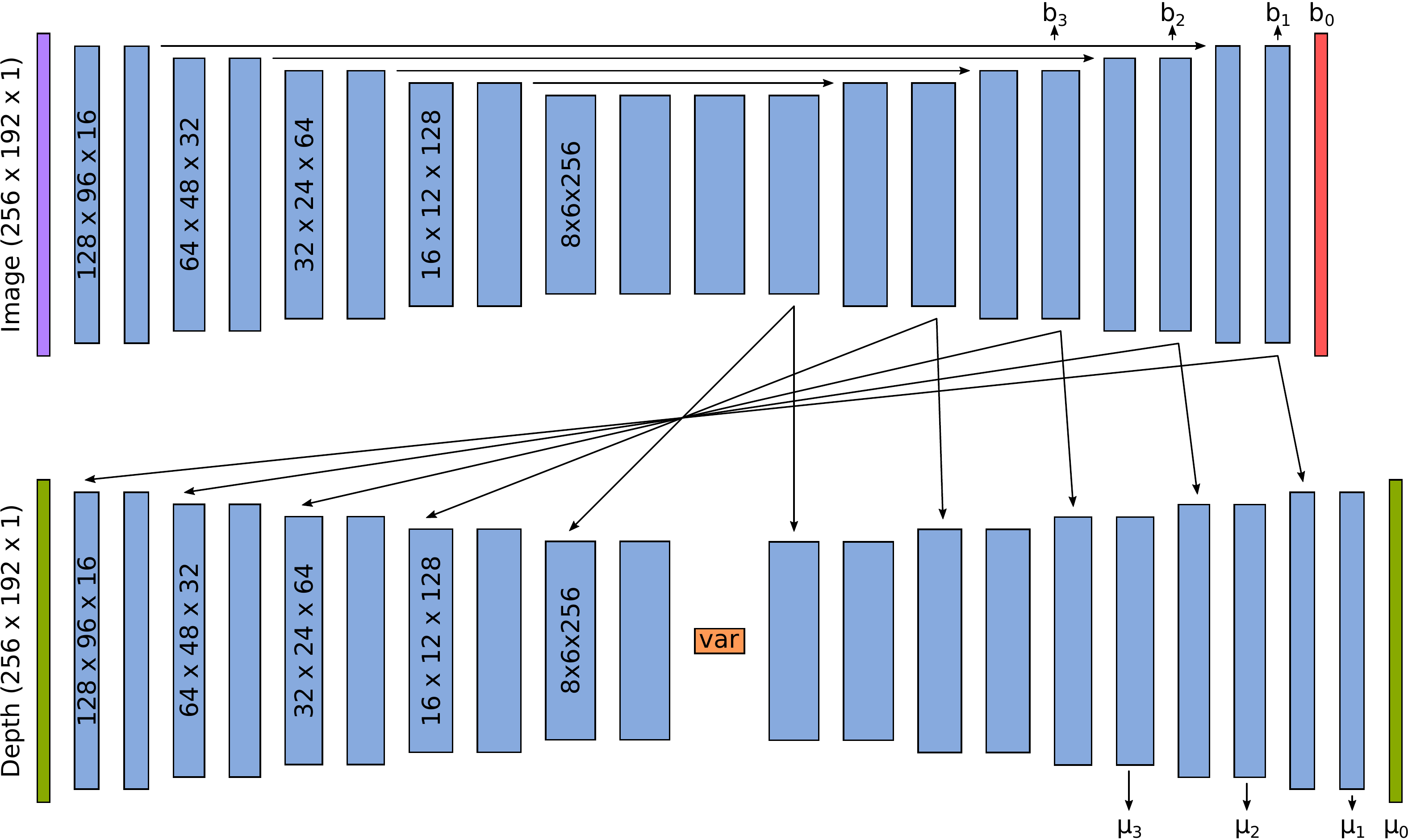}
  \end{center}
  \caption{
    Network architecture of the variational depth auto-encoder conditioned on image intensities.
    We use a U-Net to decompose the intensity image into convolutional features (the upper part of the figure).
    These features are then fed into the depth auto-encoder by concatenating them after the corresponding convolutions (denoted by arrows).
    Down-sampling is achieved by varying stride of the convolutions, while up-sampling uses bilinear interpolation (except for the last layer which uses a deconvolution).
    A variational component in the bottleneck of the depth auto-encoder is composed of two fully connected layers (512 output channels each) followed by the computation of the mean and variance, from which the latent space is then sampled.
    The network outputs the predicted mean $\mu$ and uncertainty $b$ of the depth on four pyramid levels.
   }
  \label{fig:net_det}
  \vspace{2mm}\hrule
\end{figure}

An overview of the network architecture is provided in \Cref{fig:net_det}.
The top part illustrates the U-Net \cite{Ronneberger:etal:MICCAI2015} applied on the intensity image, which first computes an increasingly coarse but high-dimensional feature representation of the input image.
This is followed by an up-sampling part with skip-layers.
The computed intensity features are then used to encode and decode the depth in the lower part of the figure.
This part is a fairly standard variational auto-encoder architecture with again a down-sampling part and an up-sampling part.
Embedded in the middle are two fully connected layers as well as the variational part, which samples the code from a Gaussian distribution and is subject to a regularisation cost (KL-divergence, see~\cite{Kingma:Welling:ICLR2014}).
The conditioning of the auto-encoder is achieved by simply concatenating the intensity features of the corresponding resolution.

Instead of predicting just raw depth values, we predict a mean $\mu$ and an uncertainty $b$ for every depth pixel.
The uncertainty is predicted from intensity only and thus is not directly influenced by the code.
Subsequently, we derive a cost term by evaluating the negative log-likelihood of the observed depth $\tilde{d}$.
This allows the network to attenuate the cost of difficult regions and to focus on reconstructing parts which can be well explained.
At test time, the learned uncertainties can also serve to gauge the reliability of the reconstruction.
In the present work we employ a Laplace distribution which has heavier tails than the traditional Gaussian distribution:
\begin{align}
  p(\tilde{d} | \mu, b)) = \frac{1}{2b} \exp\left(-\frac{|\tilde{d} - \mu|}{b}\right)~.
\end{align}
Discarding a constant offset, the negative log-likelihood thus becomes:
\begin{align}
 -\log(p(\tilde{d} | \mu, b)) = \frac{|\tilde{d} - \mu|}{b} + \log(b)~.
\end{align}
Intuitively, the network will tune the pixel-wise uncertainty $b$ such that it best attenuates the reconstruction error $|\tilde{d} - \mu|$ while being subject to a regularisation term $\log(b)$.
Using likelihoods as cost terms is a well-established method and has previously been applied to deep learning problems in computer vision~\cite{Kendall:Gal:NIPS2017,Clark:etal:CVPR2017}.

In analogy to previous work, we evaluate the error at multiple resolutions~\cite{Zhou:etal:CVPR2017}.
To this end, we create a depth image pyramid with four levels and derive the negative log-likelihood for every pixel at every level.
We increase the weight on every level by a factor of 4 in order to account for the lower pixel count.
Except for the computation of the latent distribution and the output channels, the activations are all set to ReLu.
Furthermore, for allowing pre-computation of the Jacobians (see \Cref{sec:sfm}), we explore identity activations for the depth decoder.
However, in order to retain an influence from image to code-Jacobian, we add the element-wise multiplication of every concatenation to the concatenation itself.
I.e., we increment every concatenation $[L1, L2]$ of layers $L1$ and $L2$ to $[L1, L2, L1 \odot L2]$.

\subsection{Training Setup}
The depth values of the dataset are transformed to the range $[0, 1]$.
We do this by employing a hybrid depth parametrisation which we call \emph{proximity}:
\begin{align}
  p = \frac{a}{d + a}~.
\end{align}
Given an average depth value $a$, it maps the depth in $[0, a]$ to $[0.5, 1.0]$ (similar to regular depth) and maps the depths in $[a, \infty]$ to $[0, 0.5]$ (similar to inverse depth).
This parametrisation is differentiable and better relates to the actual observable quantity (see inverse depth parametrisation~\cite{Montiel:etal:RSS2006}).

The network is trained on the SceneNet RGB-D dataset~\cite{McCormac:etal:ICCV2017} which is composed of photorealistic renderings of randomised indoor scenes.
It provides colour and depth images as well as semantic labeling and poses, out of which we only make use of the two former ones.
We make use of the ADAM optimiser~\cite{Kingma:Ba:ICLR2015} with an initial learning rate of $10^{-4}$.
We train the network for $6$ epochs while reducing the learning-rate to $10^{-6}$.

\section{Dense Warping}\label{sec:warping}
Due to the latent cost of the variational auto-encoder, the zero code can be used to obtain a likely single view depth prediction $D(I, 0)$ (see \Cref{fig:uncertainty}).
However, if overlapping views are available we can leverage stereopsis to refine the depth estimates.
This can be done by computing dense correspondences between the views:
Given the image $I_A$ and the estimated code $\b{c}_A$ of a view $A$, as well as the relative transformation $\b{T}_A^B = (\b{R}_A^B, {}_B\b{t}_A^B) \in SO(3) \times \mathbb{R}^3$ to a view $B$, we compute the correspondence for every pixel $\b{u}$ with:
\begin{align}
  w(\b{u}, \b{c}_A, \b{T}_A^B) = \pi(\b{R}_A^B \, \pi^{-1}(\b{u}, D_A[\b{u}]) + {}_B\b{t}_A^B)~,
\end{align}
where $\pi$ and $\pi^{-1}$ are the projection and inverse projection operators.
We use the shortcut $D_A = D(I_A, \b{c}_A)$ and use square brackets to denote pixel lookup.
If applied to intensity images we can for instance derive the following photometric error:
\begin{align}
  I_A[\b{u}] - I_B[w(\b{u}, \b{c}_A, \b{T}_A^B)]~.
\end{align}
The above expressions are differentiable w.r.t.\ to their inputs and we can compute the corresponding Jacobians using the chain rule:
\begin{align}
  \frac{\partial I_B[\b{v}]}{\partial {}_B\b{t}_A^B} =&~ \frac{\partial I_B[\b{v}]}{\partial \b{v}} \frac{\partial \pi(\b{x})}{\partial \b{x}} ~, \label{eq:jac_pos} \\
  \frac{\partial I_B[\b{v}]}{\partial \b{R}_A^B} =& \frac{\partial I_B[\b{v}]}{\partial \b{v}} \frac{\partial \pi(\b{x})}{\partial \b{x}} (-\b{R}_A^B \, \pi^{-1}(\b{u}, d))^{\times} ~, \\
  \frac{\partial I_B[\b{v}]}{\partial \b{c}_a} =&~ \frac{\partial I_B[\b{v}]}{\partial \b{v}} \frac{\partial \pi(\b{x})}{\partial \b{x}} \b{R}_A^B \frac{\partial \pi^{-1}(\b{u}, d)}{\partial d} \frac{\partial D_A[\b{u}]}{\partial \b{c}_A} ~, \label{eq:jac_cde}
\end{align}
where ${}^\times$ refers to the skew symmetric matrix of a 3D vector and with the abbreviations:
\begin{align}
  \b{v} =&~ w(\b{u}, \b{c}_A, \b{T}_A^B) ~, \\
  \b{x} =&~ \b{R}_A^B \, \pi^{-1}(\b{u}, D_A[\b{u}]) + {}_B\b{t}_A^B ~, \\
  d =&~ D(I_A, \b{c}_A)[\b{u}].
\end{align}

Most partial derivatives involved in \Crefrange{eq:jac_pos}{eq:jac_cde} are relatively well-known from dense tracking literature \cite{Kerl:etal:ICRA2013} and include the image gradient ($\partial I_B[\b{v}] / \partial \b{v}$), the differential of the projection ($\partial \pi(\b{x}) / \partial \b{x}$), as well as transformation related derivatives (also refer to \cite{Bloesch:etal:CoRR2016} for more details).
The last term in \Cref{eq:jac_cde}, $\partial D_A[\b{u}] / \partial \b{c}_A$, is the derivative of the depth w.r.t. the code.
Since it involves many convolutions, it is computationally costly to evaluate (up to \unit[1]{sec} depending on the size of the network).
In case of a linear decoder this term can be pre-computed which significantly accelerates the evaluation of the Jacobians.

\section{Inference Framework}
\subsection{N-Frame Structure from Motion (Mapping)}\label{sec:sfm}
\begin{figure}[t]
  \begin{center}
    \includegraphics[width=1.0\linewidth]{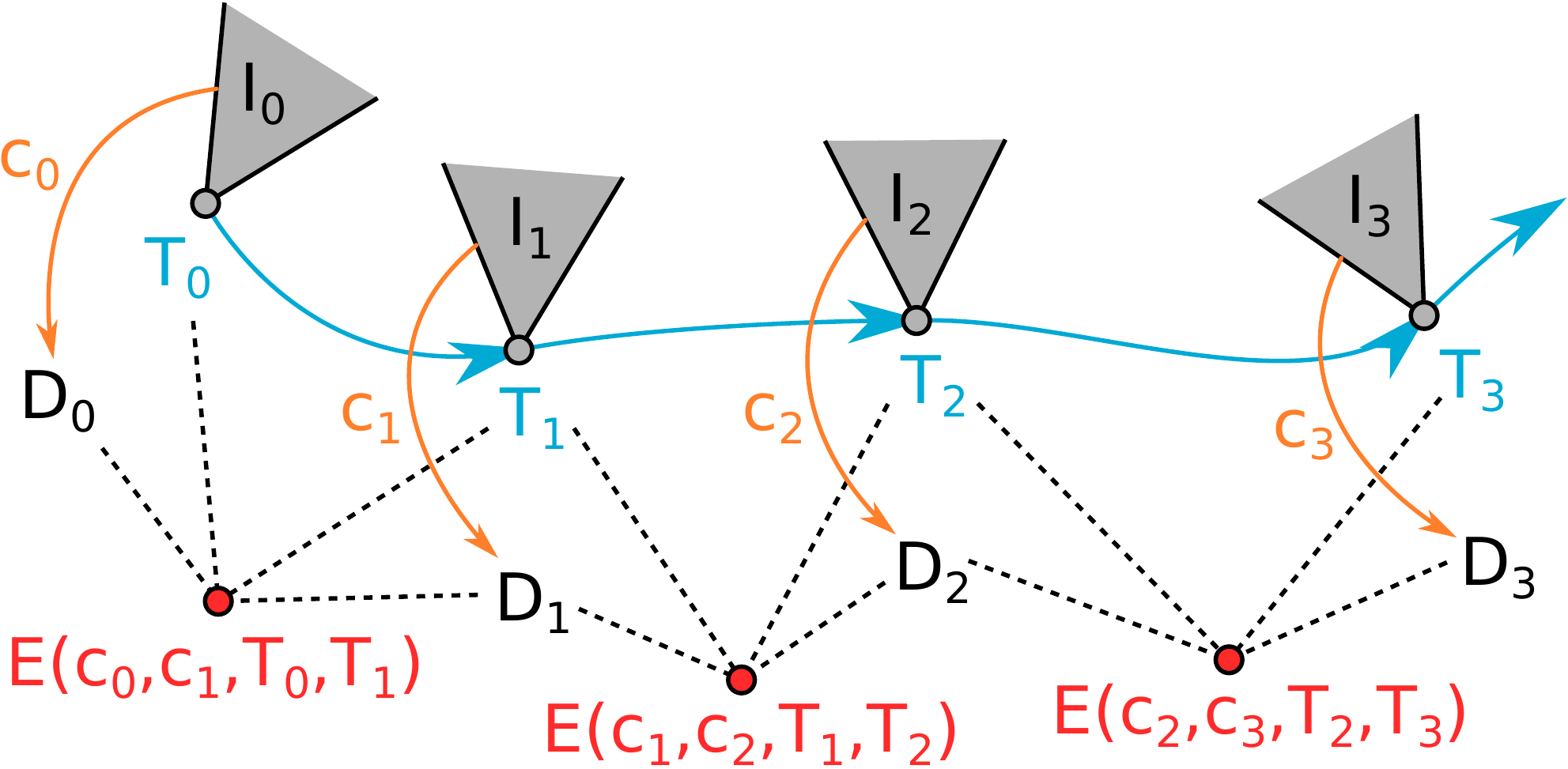}
  \end{center}
  \caption{
    Illustration of the \ac{SfM} system.
    The image $I_i$ and corresponding code $\b{c}_i$ in each frame are used to estimate the depth $D_i$.
    Given estimated poses $\b{T}_i$, we derive relative error terms between the frames (photometric and geometric).
    We then jointly optimise for geometry ($\b{c}_i$) and motion ($\b{T}_i$) by using a standard second-order method.
  }
  \label{fig:factor_graph}
  \vspace{2mm}\hrule
\end{figure}

The proposed depth parametrisation is used to construct a dense $N$-frame \ac{SfM} framework (see \Cref{fig:factor_graph}).
We do this by assigning an unknown code and an unknown pose to every frame.
All codes and poses are initialised to zero and identity, respectively.
For two frames $A$ and $B$ with overlapping field of view we then derive photometric and geometric residuals, $E_\mathrm{pho}$ and $E_\mathrm{geo}$, as follows:
\begin{align}
  E_\mathrm{pho} = L_p\big(&I_A[\b{u}] - I_B[w(\b{u}, \b{c}_A, \b{T}_A^B)]\big)~, \\
  E_\mathrm{geo} = L_g\big(&D_A[\b{u}] - D_B[w(\b{u}, \b{c}_A, \b{T}_A^B)]\big)~.
\end{align}
The loss functions $L_\mathrm{pho}$ and $L_\mathrm{geo}$ have the following masking and weighting functionality:
(i) mask invalid correspondences, (ii) apply relative weighting to geometric and photometric errors, (iii) apply a Huber weighting, (iv) down-weight errors on strongly slanted surfaces, and (v) down-weight pixels which might be occluded (only $L_\mathrm{pho}$).

In order to optimise both sets of residuals w.r.t.\ our motion and geometry we compute the Jacobians w.r.t.\ all codes and poses according to \Cref{sec:warping}.
As mentioned above, we investigate the applicability of linear decoding networks (see \Cref{sec:det_net}) as this allows us to compute the Jacobian of the decoder $D(I, \b{c})$ w.r.t.\ the code $\b{c}$ only once per keyframe.
After computing all residuals and Jacobians we apply a damped Gauss-Newton algorithm in order to find the optimal codes and poses of all frames.

\subsection{Tracking (Localisation)}
The tracking system, responsible for estimating the pose of keyframes with respect to an existing keyframe map, can be built much in the spirit of the above \ac{SfM} approach.
The current frame is paired with the last keyframe and the estimated relative pose results from a cost-minimisation problem.
In our vision-only setup we do not have access to the current depth image (except for a rough guess), and thus in contrast to the described \ac{SfM} system we do not integrate a geometric cost.

In order to increase tracking robustness we perform a coarse to fine optimisation by first doing the dense alignment on the low depth image resolutions.

\subsection{SLAM System}
We implement a preliminary system for Simultaneous Localisation and Mapping inspired by PTAM~\cite{Klein:Murray:ISMAR2007}
where we alternate between tracking and mapping.
The initialisation procedure takes two images and jointly optimises for their relative pose and the codes of each frame.
After that we can track the current camera pose w.r.t.\ the last keyframe.
Once a certain baseline is achieved we add a keyframe to the map and perform a global optimisation, before continuing with the tracking.
If the maximum number of keyframes is reached we marginalise old keyframes and thereby obtain a linear prior on the remaining keyframes.
In a 4-keyframes setup, we achieve a map update rate of 5 Hz, which if we do not have to add keyframes too frequently is enough for real-time performance.
The system currently relies on Tensorflow for image warping, and could be sped up with a more targeted warping and optimisation system which are both part of future work.

\section{Experimental Evaluation and Discussion}

Please also see our submitted video which includes demonstrations of our results and system \url{http://www.imperial.ac.uk/dyson-robotics-lab/projects/codeslam/}.

\subsection{Image Conditioned Depth Encoding}

\begin{figure}[t]
  \begin{center}
    \includegraphics[width=1.0\linewidth]{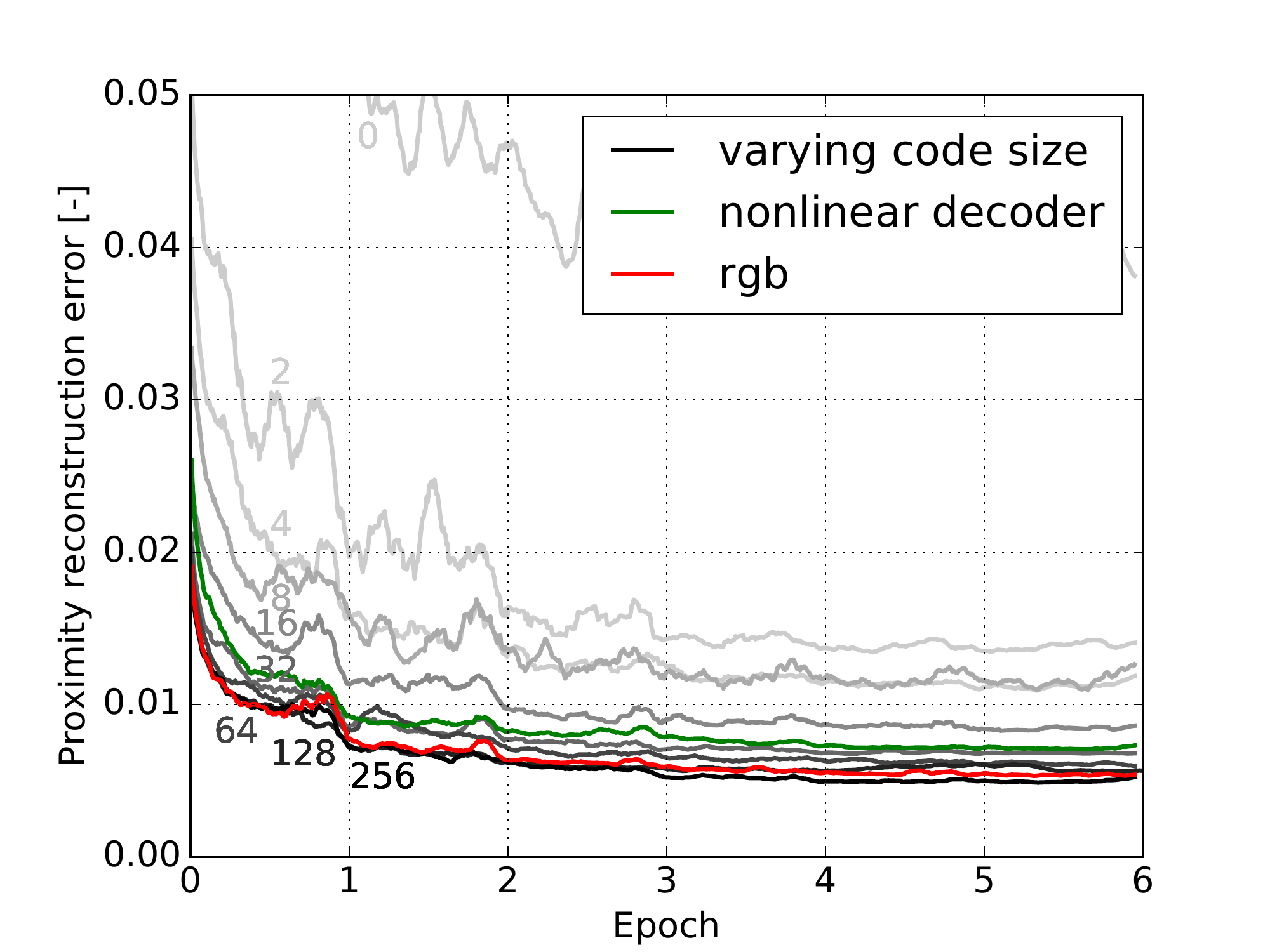}
  \end{center}
  \caption{
    Validation loss during training on the per-pixel proximity errors.
    As the reference implementation, we use a network trained on greyscale images with a linear decoder.
    Lower losses can be achieved by increasing the code size (increasing shades of grey).
    Using a nonlinear decoder or colour images during training does not affect the results in a significant way.
  }
  \label{fig:training_error}
  \vspace{2mm}\hrule
\end{figure}

\begin{figure}[t]
  \begin{center}
    \small
    \rotatebox{90}{\hspace{5.5mm} Uncertainty \hspace{12mm} Zero code \hspace{13mm} Input image} \includegraphics[width=0.90\linewidth]{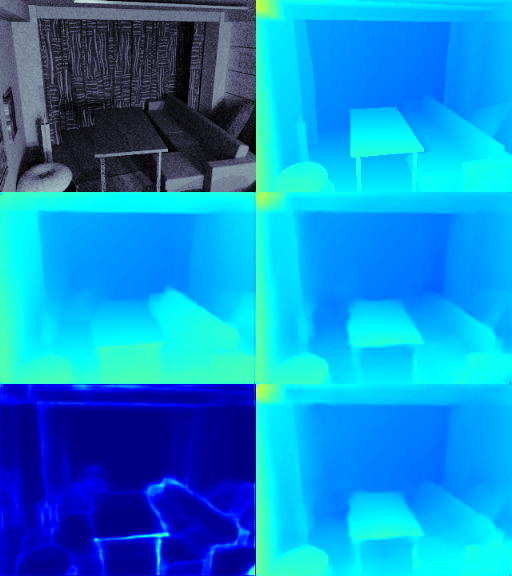}
    \rotatebox{90}{\hspace{2mm} Optimised depth \hspace{7mm} Encoded depth \hspace{10mm} Groundtruth}
  \end{center}
  \caption{
    An example image passed through encoding and decoding.
    Top left: input image.
    Top right: ground truth depth.
    Middle left: zero code reconstruction (image only prediction).
    Middle right: decoded depth (code from encoder).
    Bottom left: estimated reconstruction uncertainty (scaled four times for visibility).
    Bottom right: optimised depth (code minimising reconstruction error).
  }
  \label{fig:uncertainty}
  \vspace{2mm}\hrule
\end{figure}

\begin{figure}[t]
  \begin{center}
    \small
    \rotatebox{90}{\hspace{1mm} Reconstr. \hspace{0.5mm} Groundtr.}
    \includegraphics[width=0.239\linewidth,trim={257px 193px 0 0},clip]{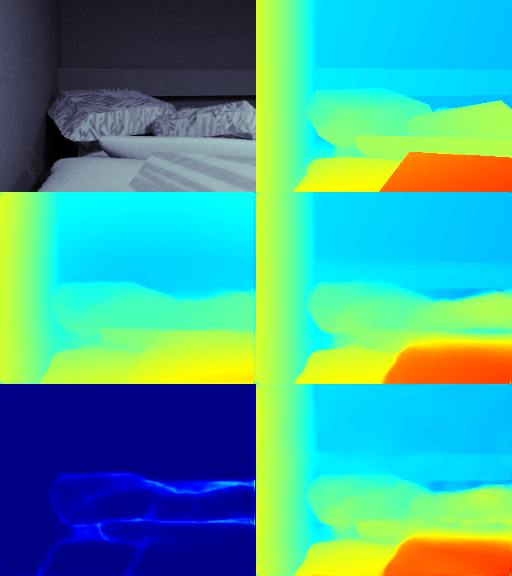}%
    \includegraphics[width=0.239\linewidth,trim={257px 193px 0 0},clip]{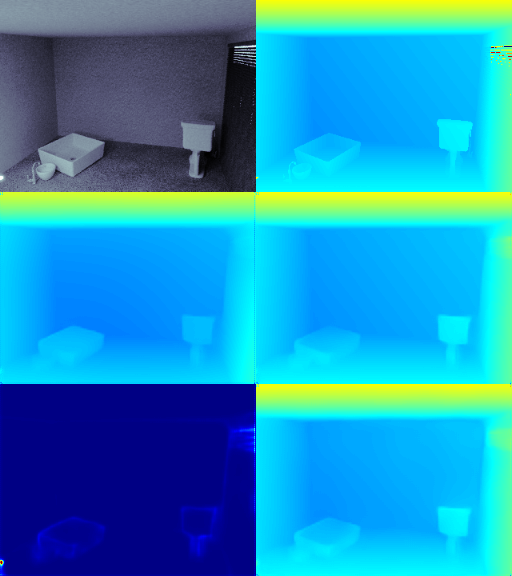}%
    \includegraphics[width=0.239\linewidth,trim={257px 193px 0 0},clip]{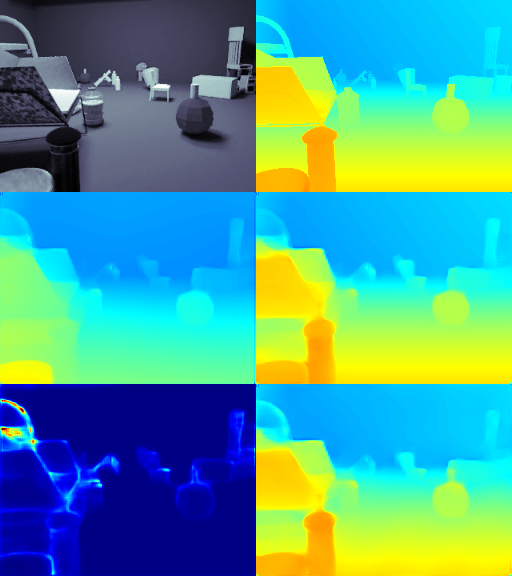}%
    \includegraphics[width=0.239\linewidth,trim={257px 193px 0 0},clip]{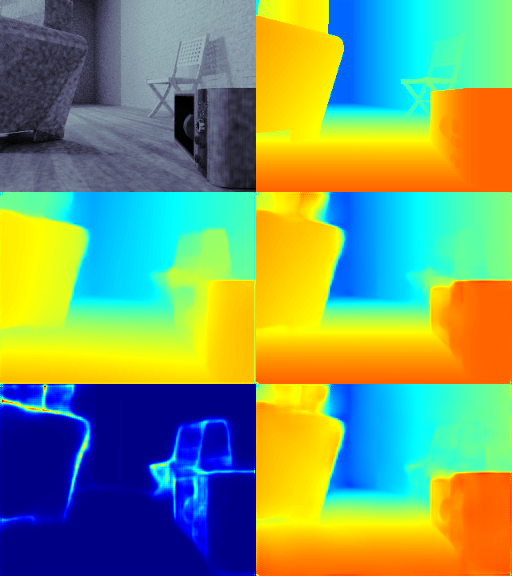}
  \end{center}
  \caption{
    Encodings of different depth images.
    The encoding allows to capture even fine geometrical details.
  }
  \label{fig:scenenet_encodings}
  \vspace{2mm}\hrule
\end{figure}

\begin{figure}[t]
  \begin{center}
    \small
    \hspace{4mm} Entry 1 \hspace{14mm} Entry 2 \hspace{14mm} Entry 3 \\
    \rotatebox{90}{\hspace{3.5mm} Image 2 \hspace{6.5mm} Image 1}
    \includegraphics[width=0.9\linewidth]{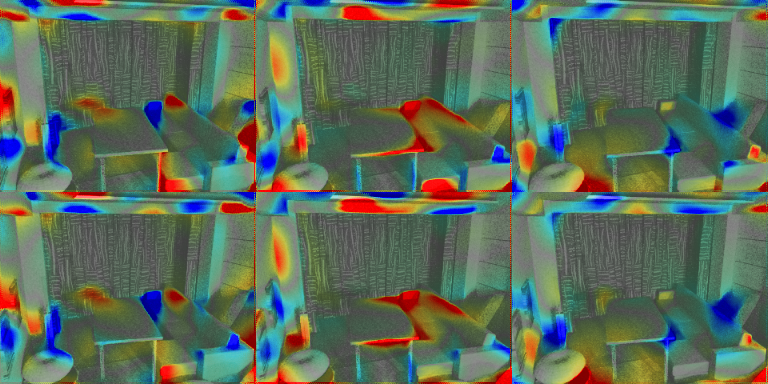}
  \end{center}
  \caption{
    Visualisation of the influence of the code on depth reconstruction.
    The Jacobian of the depth w.r.t. a specific code entry is used to colourise the input image (blue and red depict negative and positive values, respectively).
    Columns represent code entries (1-3).
    Rows represent two different input images.
  }
  \label{fig:jacobians}
  \vspace{2mm}\hrule
\end{figure}

First we present results and insights related to our key concept of encoding depth maps conditioned on intensity images.

We trained and compared multiple variations of our network.
Our reference network has a code size of 128, employs greyscale image information only, and makes use of a linear decoder network in order to speed up Jacobian computation.
\Cref{fig:training_error} shows results on reconstruction accuracy using different code sizes as well as setups with RGB information and nonlinear depth decoding.
The use of colour or nonlinear decoding did not significantly affect the accuracy.
With regard to code size, we observe a saturation of the accuracy at a code size of 128; there is little to be gained from making the code bigger.
This value may be surprisingly low, but the size seems to be large enough to transmit the information that can be captured in the code by the proposed network architecture.

\Crefrange{fig:uncertainty}{fig:jacobians} provide some insight into how our image conditioned depth encoding works.
In \Cref{fig:uncertainty} we show how we encode a depth image into a code of size 128.
Using the corresponding intensity image this can then be decoded into a reconstructed depth image, which captures all of the main scene elements well.
We also show the reconstruction when passing a zero code to the decoder as well as with a code that is optimised for minimal reconstruction error.
The zero code captures some of the geometrical details but fails to properly reconstruct the entire scene.
The reconstruction with the optimised code is very similar to the one with the code from the encoder which indicates that the encoder part of the network works well.
The associated depth uncertainty is also visualised and exhibits higher magnitudes in the vicinity of depth discontinuities and around shiny image regions (but not necessarily around high image gradients in general).
Further examples of depth encoding are shown in \Cref{fig:scenenet_encodings}.

In \Cref{fig:jacobians} we visualise the Jacobians of the depth image w.r.t. to the code entries.
An interesting observation is that the single code entries seem to correspond to specific image regions and, to some extent, respect boundaries given by the intensity image.
While the regions seem to be slightly fragmented, the final reconstructions will always be a linear combination of the effect of all code entries.
We also compare the regions of influence for two different but similar images and can observe a certain degree of consistency.

\subsection{Structure from Motion}

\begin{figure*}[t]
  \begin{center}
    \includegraphics[width=1.0\linewidth]{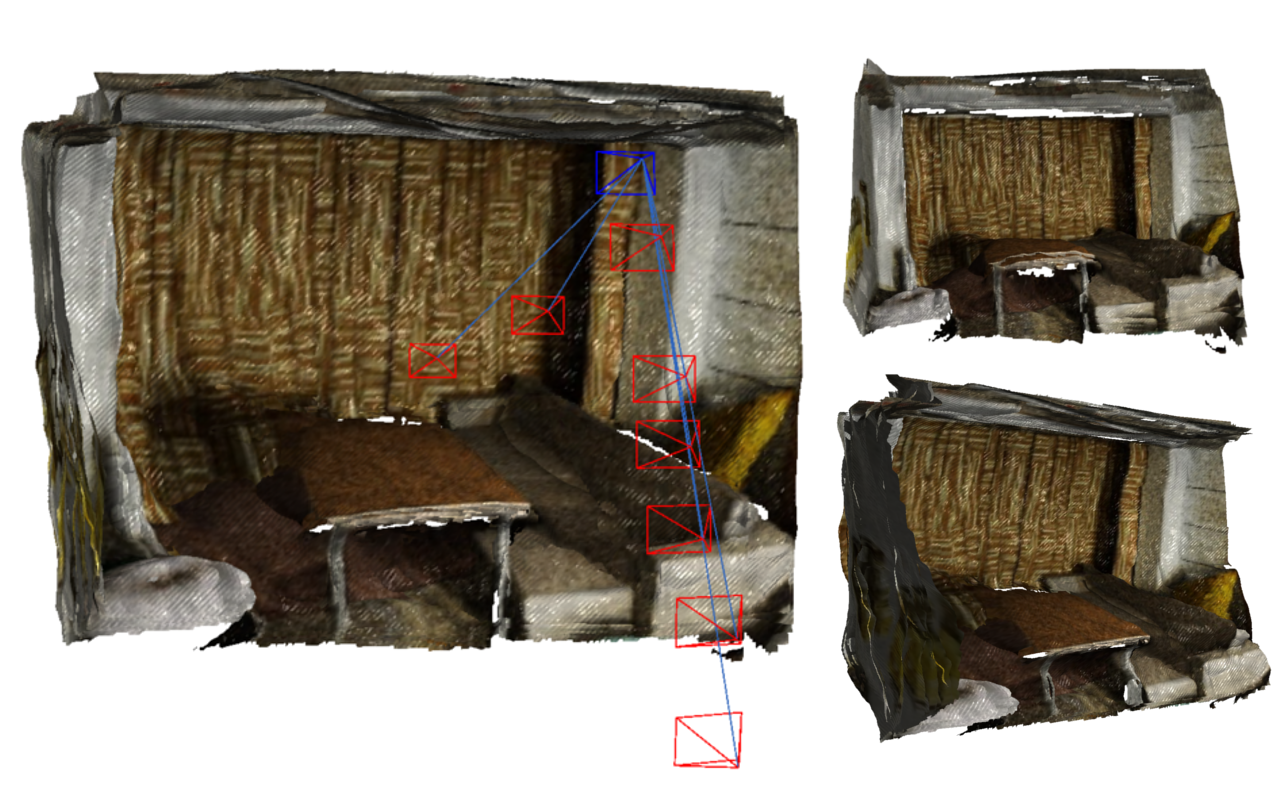}
  \end{center}
  \caption{
    Monocular 3D reconstruction using 9 keyframes.
    During optimisation a selected master keyframe is paired with the other frames.
    The depth images of all frames are used for the 3D rendering.
    The employed geometric error term ensures the consistency between the depth of the different views.
  }
  \label{fig:reconstruction_scenenet}
  \vspace{2mm}\hrule
\end{figure*}

\begin{table}
  \def\arraystretch{1.0}
  \setlength{\tabcolsep}{0.5em}
  \begin{tabular}{|c|c|c|c|c|c|c|}
    \hline 
    \# frames & 1 & 2 & 3 & 4 & 5 & 6 \\ 
    \hline
    RMSE [$10^{-2}$] & 2.65 & 2.47 & 2.31 & 2.39 & 
    2.30 & 2.14 \\ 
    \hline 
  \end{tabular}
  \vspace{0.1cm}
  \caption{
    RMS of pixel proximity estimation error with different amounts of master keyframe-frame pairs in the optimisation problem.
    The error is evaluated between the master keyframe proximity and its corresponding ground truth proximity.
    Frames 1-3: downward-backwards motion.
    Frames 4-6: left-forward motion.
  }
  \label{tab:reconstruction_scenenet}
  \vspace{2mm}\hrule
\end{table}

\begin{figure}[t]
  \begin{center}
    \small
    \rotatebox{90}{\hspace{0.5mm} Reconstr. \hspace{2.5mm} Image}
    \includegraphics[width=0.95\linewidth]{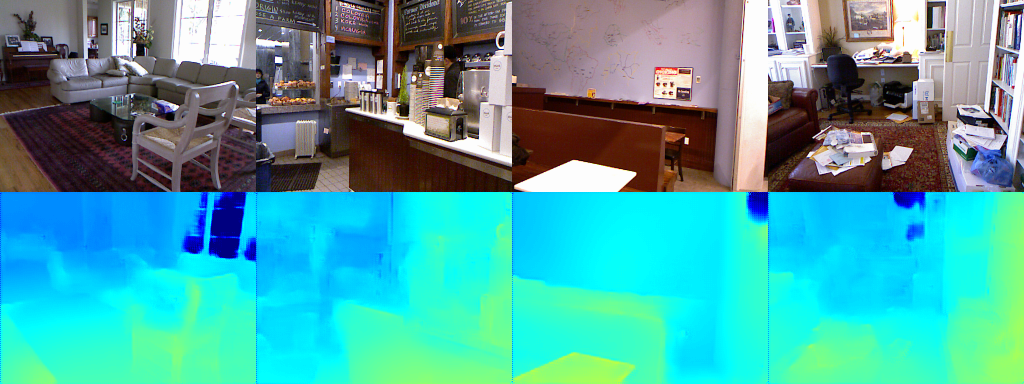}
  \end{center}
  \caption{
    Two-frame \ac{SfM} on selected pairs from the NYU V2 dataset.
    Top row presents one of the images used for reconstruction, while the bottom row contains respective depth estimates.
    The main elements of all scenes can be well perceived in the depth image.
    The overexposed image regions saturate to infinite depth values, which is a result of using the SceneNet RGB-D dataset for training, which contains many scenes with windows (similar to the one in the left image).
  }
  \label{fig:live_two_frame_reconstructions}
  \vspace{2mm}\hrule
\end{figure}

The proposed low dimensional encoding enables continuous refinement of the depth estimates as more overlapping keyframes are integrated.
In order to test this, we have implemented an \ac{SfM} system which incrementally pairs one pre-selected frame with all the remaining frames (which were selected from SceneNet RGB-D).
\Cref{tab:reconstruction_scenenet} shows the obtained reconstruction error w.r.t. the number of frames that are connected to the first frame.
The observed reduction of the reconstruction error well illustrates the strength of the employed probabilistic inference method, application of which is enabled by the low dimensionality of the optimisation space.
The magnitude of depth refinement depends on the information content of the new frames (whether they present the scene under a new view and exhibit sufficient baseline).
\Cref{fig:reconstruction_scenenet} presents a 3D reconstruction based on 9 frames for the scene used in the above error computations.
Since in this rendering all the frame depth maps are superimposed, one can observe the quality of the alignment.
In a future full SLAM system, these keyframes would be fused together in order to form a single global scene.
Before visualisation, high frequency elements are removed from the depth maps with bilateral filtering and highly slanted mesh elements are cropped.

Being exposed to a large variety of depth images during training, the proposed network embeds geometry priors in its weights.
These learned priors seem to generalise to real scenes as well:
\Cref{fig:sfm2_euroc} depicts a two-frame reconstruction with images from the real image EuRoC dataset~\cite{Burri:etal:IJRR2016} taken by a drone in an industrial setting.
The result corresponds to 50 optimisation steps, each taking around \unit[100]{ms} to complete.
Since significant exposure changes occur between the images, we perform an affine illumination correction of the frames.
The validation of the two-frame reconstruction performance is of high importance as it is directly connected to the initialisation procedure of the full SLAM system.
In order to further highlight its effectiveness we include results on a selection of pairs taken from the NYU V2 dataset~\cite{Silberman:etal:ECCV2012} (\Cref{fig:live_two_frame_reconstructions}).
 
\subsection{SLAM System}

\begin{figure}[t]
  \begin{center}
    \includegraphics[width=1.0\linewidth]{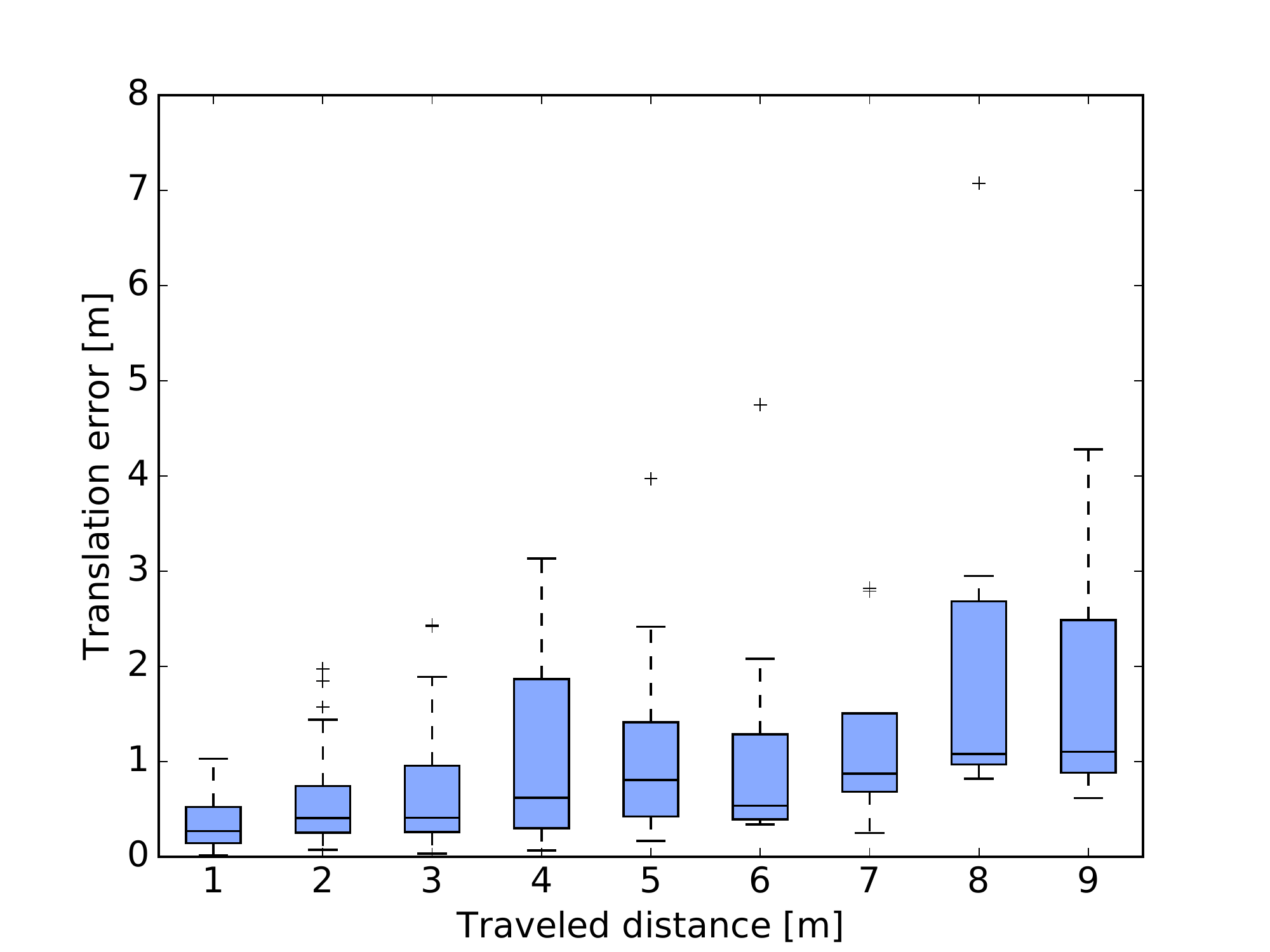}
  \end{center}
  \caption{
    Translation error versus traveled distance on the EuRoC dataset MH02.
    Despite training the auto-encoder on SceneNet\ RGB-D, its decoder generalises to other datasets (after correcting for camera intrinsics).
  }
  \label{fig:slam_traj}
  \vspace{2mm}\hrule
\end{figure}

\begin{figure}[t]
  \begin{center}
    \small
    \hspace{3mm} Image \hspace{12mm} Reconstruction \hspace{10mm} Shading \\[1mm]
    \rotatebox{90}{\hspace{1.5mm} Keyframe 4 \hspace{3.5mm} Keyframe 1}
    \includegraphics[width=0.95\linewidth]{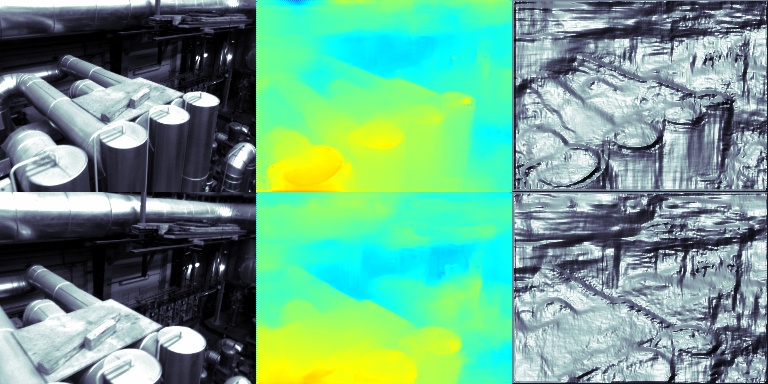}
  \end{center}
  \caption{
    Example structure from motion results on frames from the EuRoC dataset.
    From the left: image, estimated proximity, shaded proximity.
  }
  \label{fig:MH02_depth_fl}
  \vspace{2mm}\hrule
\end{figure}

In contrast to most dense approaches, our low dimensional geometry encoding allows joint optimisation of motion and geometry.
Furthermore, due to the inherent prior contained in the encoding, the framework is able to deal with rotational motions only.
The system is tested in a sliding window visual odometry mode on the EuRoC dataset on trajectory MH\_02\_easy.
Even though the dataset is significantly different from the data the network is trained on (with many metallic parts and many reflections), the proposed system is able to run through most of this arguably very difficult dataset (we do not use the available IMU data).

\Cref{fig:slam_traj} shows the error against traveled distance.
While this cannot compete with a state-of-the art visual-inertial system, it performs respectably for a vision only-system and exhibits an error of roughly \unit[1]{m} for a traveled distance of \unit[9]{m}.
In \Cref{fig:MH02_depth_fl} the first and last key-frame of our 4-frame sliding window system are illustrated.
This shows the intensity image of the encountered scene together with the estimated proximity image and a normal based shading.
Considering that the network was trained on artificial images only which were very different in their nature, the reconstructed depth is sensible and allows for reliable camera tracking.

\section{Conclusions}

We have shown that a learned representation for depth which is conditioned on image data provides an important advance towards future SLAM systems.
By employing an auto-encoder like training setup, the proposed representation can contain generic and detailed dense scene information while allowing efficient probabilistic joint optimisation together with camera poses.

In near future work, we will use the components demonstrated here to build a full real-time keyframe-based SLAM system.
Learned visual motion estimation methods could surely be brought in here as priors for robust tracking.
In addition to that, the training of the network should be extended in order to include real data as well.
This could be done by using an RGB-D dataset, but might also be achieved with intensity information only in an self-supervised manner, based on photometric error as loss.

In the longer term, we would like to move beyond a keyframe-based approach, where our dense geometry representations are tied to single images, and work on learned but optimisable compact representations for general 3D geometry, eventually tying our work up with 3D object recognition.

\section{Acknowledgements}

Research presented in this paper has been supported by Dyson Technology Ltd.

{\small
\bibliographystyle{ieee}
\bibliography{robotvision}

\begin{thebibliography}{10}\itemsep=-1pt

\bibitem{Bloesch:etal:CoRR2016}
M.~Bloesch, H.~Sommer, T.~Laidlow, M.~Burri, G.~N{\"{u}}tzi, P.~Fankhauser,
  D.~Bellicoso, C.~Gehring, S.~Leutenegger, M.~Hutter, and R.~Siegwart.
\newblock {A Primer on the Differential Calculus of 3D Orientations}.
\newblock {\em CoRR}, abs/1606.0, 2016.

\bibitem{Burri:etal:IJRR2016}
M.~Burri, J.~Nikolic, P.~Gohl, T.~Schneider, J.~Rehder, S.~Omari, M.~W.
  Achtelik, and R.~Siegwart.
\newblock {The EuRoC Micro Aerial Vehicle Datasets}.
\newblock {\em {International Journal of Robotics Research ({IJRR})}},
  35(10):1157--1163, September 2016.

\bibitem{Cadena:etal:RSS2016}
C.~Cadena, A.~Dick, and I.~D. Reid.
\newblock {Multi-modal Auto-Encoders as Joint Estimators for Robotics Scene
  Understanding}.
\newblock In {\em {Proceedings of Robotics: Science and Systems ({RSS})}},
  2016.

\bibitem{Clark:etal:CVPR2017}
R.~Clark, S.~Wang, H.~Wen, A.~Markham, and N.~Trigoni.
\newblock {VidLoc}: A deep spatio-temporal model for 6-dof video-clip
  relocalization.
\newblock In {\em {Proceedings of the {IEEE} Conference on Computer Vision and
  Pattern Recognition ({CVPR})}}, 2017.

\bibitem{Clark:etal:AAAI2017}
R.~Clark, S.~Wang, H.~Wen, A.~Markham, and N.~Trigoni.
\newblock {VINet}: Visual-inertial odometry as a sequence-to-sequence learning
  problem.
\newblock In {\em {Proceedings of the National Conference on Artificial
  Intelligence ({AAAI})}}, 2017.

\bibitem{Davison:ICCV2003}
A.~J. Davison.
\newblock {Real-Time Simultaneous Localisation and Mapping with a Single
  Camera}.
\newblock In {\em {Proceedings of the International Conference on Computer
  Vision ({ICCV})}}, 2003.

\bibitem{Durrant-Whyte:etal:RAM2006}
H.~Durrant-Whyte and T.~Bailey.
\newblock {Simultaneous Localisation and Mapping ({SLAM}): Part {I} The
  Essential Algorithms}.
\newblock {\em {IEEE} Robotics and Automation Magazine}, 13(2):99--110, 2006.

\bibitem{Eigen:etal:NIPS2014}
D.~Eigen, C.~Puhrsch, and R.~Fergus.
\newblock {Depth Map Prediction from a Single Image using a Multi-Scale Deep
  Network}.
\newblock In {\em {Neural Information Processing Systems ({NIPS})}}, 2014.

\bibitem{Engel:etal:PAMI2017}
J.~Engel, V.~Koltun, and D.~Cremers.
\newblock Direct sparse odometry.
\newblock {\em {{IEEE} Transactions on Pattern Analysis and Machine
  Intelligence ({PAMI})}}, 2017.

\bibitem{Engel:etal:ECCV2014}
J.~Engel, T.~Schoeps, and D.~Cremers.
\newblock {LSD-SLAM}: Large-scale direct monocular {SLAM}.
\newblock In {\em {Proceedings of the European Conference on Computer Vision
  ({ECCV})}}, 2014.

\bibitem{Garg:etal:ECCV2016}
R.~Garg, V.~K.~B. G, G.~Carneiro, and I.~Reid.
\newblock Unsupervised {CNN} for single view depth estimation: Geometry to the
  rescue.
\newblock In {\em {Proceedings of the European Conference on Computer Vision
  ({ECCV})}}, 2016.

\bibitem{Kaess:ICRA2015}
M.~Kaess.
\newblock Simultaneous localization and mapping with infinite planes.
\newblock In {\em {Proceedings of the {IEEE} International Conference on
  Robotics and Automation ({ICRA})}}, 2015.

\bibitem{Kendall:Gal:NIPS2017}
A.~Kendall and Y.~Gal.
\newblock What uncertainties do we need in bayesian deep learning for computer
  vision?
\newblock In {\em {Neural Information Processing Systems ({NIPS})}}, 2017.

\bibitem{Kerl:etal:ICRA2013}
C.~Kerl, J.~Sturm, and D.~Cremers.
\newblock Robust odometry estimation for {RGB-D} cameras.
\newblock In {\em {Proceedings of the {IEEE} International Conference on
  Robotics and Automation ({ICRA})}}, 2013.

\bibitem{Kingma:Ba:ICLR2015}
D.~P. Kingma and J.~Ba.
\newblock Adam: {A} method for stochastic optimization.
\newblock In {\em {Proceedings of the International Conference on Learning
  Representations ({ICLR})}}, 2015.

\bibitem{Kingma:Welling:ICLR2014}
D.~P. Kingma and M.~Welling.
\newblock {Auto-Encoding Variational Bayes}.
\newblock In {\em {Proceedings of the International Conference on Learning
  Representations ({ICLR})}}, 2014.

\bibitem{Klein:Murray:ISMAR2007}
G.~Klein and D.~W. Murray.
\newblock {Parallel Tracking and Mapping for Small {AR} Workspaces}.
\newblock In {\em {Proceedings of the International Symposium on Mixed and
  Augmented Reality ({ISMAR})}}, 2007.

\bibitem{Liu:etal:2015}
F.~Liu, C.~Shen, and G.~Lin.
\newblock {Deep Convolutional Neural Fields for Depth Estimation from a Single
  Image}.
\newblock In {\em {Proceedings of the {IEEE} Conference on Computer Vision and
  Pattern Recognition ({CVPR})}}, 2015.

\bibitem{McCormac:etal:ICCV2017}
J.~McCormac, A.~Handa, S.~Leutenegger, and A.~J. Davison.
\newblock {SceneNet RGB-D}: Can {5M} synthetic images beat generic {ImageNet}
  pre-training on indoor segmentation?
\newblock In {\em {Proceedings of the International Conference on Computer
  Vision ({ICCV})}}, 2017.

\bibitem{Montiel:etal:RSS2006}
J.~M.~M. Montiel, J.~Civera, and A.~J. Davison.
\newblock {Unified Inverse Depth Parametrization for Monocular {SLAM}}.
\newblock In {\em {Proceedings of Robotics: Science and Systems ({RSS})}},
  2006.

\bibitem{Mur-Artal:etal:TRO2015}
R.~Mur-Artal, J.~M.~M. Montiel, and J.~D. Tard{\'o}s.
\newblock {{ORB-SLAM}: a Versatile and Accurate Monocular SLAM System}.
\newblock {\em {{IEEE} Transactions on Robotics ({T-RO})}}, 31(5):1147--1163,
  2015.

\bibitem{Newcombe:etal:ICCV2011}
R.~A. Newcombe, S.~Lovegrove, and A.~J. Davison.
\newblock {{DTAM}: Dense Tracking and Mapping in Real-Time}.
\newblock In {\em {Proceedings of the International Conference on Computer
  Vision ({ICCV})}}, 2011.

\bibitem{Platinsky:etal:ICRA2017}
L.~Platinsky, A.~J. Davison, and S.~Leutenegger.
\newblock Monocular visual odometry: Sparse joint optimisation or dense
  alternation?
\newblock In {\em {Proceedings of the {IEEE} International Conference on
  Robotics and Automation ({ICRA})}}, 2017.

\bibitem{Ronneberger:etal:MICCAI2015}
O.~Ronneberger, P.~Fischer, and T.~Brox.
\newblock {U-Net}: Convolutional networks for biomedical image segmentation.
\newblock In {\em {Proceedings of the International Conference on Medical Image
  Computing and Computer Assisted Intervention ({MICCAI})}}, 2015.

\bibitem{Rumelhart:1986}
D.~E. Rumelhart, G.~E. Hinton, and R.~J. Williams.
\newblock Learning internal representations by error propagation.
\newblock In D.~E. Rumelhart, J.~L. McClelland, and C.~PDP Research~Group,
  editors, {\em Parallel Distributed Processing: Explorations in the
  Microstructure of Cognition}, volume~1, pages 318--362. MIT Press, Cambridge,
  MA, USA, 1986.

\bibitem{Salas-Moreno:etal:ISMAR2014}
R.~F. Salas-Moreno, B.~Glocker, P.~H.~J. Kelly, and A.~J. Davison.
\newblock Dense planar {SLAM}.
\newblock In {\em {Proceedings of the International Symposium on Mixed and
  Augmented Reality ({ISMAR})}}, 2014.

\bibitem{Salas-Moreno:etal:CVPR2013}
R.~F. Salas-Moreno, R.~A. Newcombe, H.~Strasdat, P.~H.~J. Kelly, and A.~J.
  Davison.
\newblock {{SLAM++}: Simultaneous Localisation and Mapping at the Level of
  Objects}.
\newblock In {\em {Proceedings of the {IEEE} Conference on Computer Vision and
  Pattern Recognition ({CVPR})}}, 2013.

\bibitem{Silberman:etal:ECCV2012}
N.~Silberman, D.~Hoiem, P.~Kohli, and R.~Fergus.
\newblock Indoor segmentation and support inference from {RGBD} images.
\newblock In {\em {Proceedings of the European Conference on Computer Vision
  ({ECCV})}}, 2012.

\bibitem{Tateno:etal:CVPR2017}
K.~Tateno, F.~Tombari, I.~Laina, and N.~Navab.
\newblock {CNN-SLAM}: Real-time dense monocular slam with learned depth
  prediction.
\newblock In {\em {Proceedings of the {IEEE} Conference on Computer Vision and
  Pattern Recognition ({CVPR})}}, 2017.

\bibitem{Ummenhofer:etal:ARXIV2016}
B.~Ummenhofer, H.~Zhou, J.~Uhrig, N.~Mayer, E.~Ilg, A.~Dosovitskiy, and
  T.~Brox.
\newblock {DeMoN}: Depth and motion network for learning monocular stereo.
\newblock {\em arXiv preprint arXiv:1612:02401}, 2016.

\bibitem{Wang:etal:ICRA2017}
S.~Wang, R.~Clark, H.~Wen, and N.~Trigoni.
\newblock {DeepVO}: Towards end to end visual odometry with deep recurrent
  convolutional neural networks.
\newblock In {\em {Proceedings of the {IEEE} International Conference on
  Robotics and Automation ({ICRA})}}, 2017.

\bibitem{Weerasekera:etal:ICRA2017}
C.~S. Weerasekera, Y.~Latif, R.~Garg, and I.~Reid.
\newblock {Dense monocular reconstruction using surface normals}.
\newblock In {\em {Proceedings of the {IEEE} International Conference on
  Robotics and Automation ({ICRA})}}, 2017.

\bibitem{Yin:Shi:CVPR2018}
Z.~Yin and J.~Shi.
\newblock {GeoNet: Unsupervised Learning of Dense Depth, Optical Flow and
  Camera Pose}.
\newblock In {\em {Proceedings of the {IEEE} Conference on Computer Vision and
  Pattern Recognition ({CVPR})}}, 2018.

\bibitem{Zhou:etal:CVPR2017}
T.~Zhou, M.~Brown, N.~Snavely, and D.~G. Lowe.
\newblock Unsupervised learning of depth and ego-motion from video.
\newblock In {\em {Proceedings of the {IEEE} Conference on Computer Vision and
  Pattern Recognition ({CVPR})}}, 2017.

\end{thebibliography}
}

\end{document}